%% file: main.tex
\documentclass[letterpaper, 10 pt, conference]{ieeeconf}  

\IEEEoverridecommandlockouts                              

\usepackage{graphicx}
\usepackage{caption}
\usepackage{subcaption}
\usepackage{booktabs}
\usepackage{amsmath,amssymb,wasysym}
\usepackage{color, colortbl}
\definecolor{Gray}{gray}{0.95}
\usepackage{multirow}
\usepackage{xcolor}
\usepackage{ifthen}
\usepackage{cite} 

\include{inputs/999-inline-notes}

\newcommand{\BC}{$\mathtt{BC}$}
\newcommand{\KNN}{$\mathtt{k\text{-}NN}$}
\newcommand{\OBJC}{$\mathtt{ObjC}$}
\newcommand{\ROBOTC}{$\mathtt{RobotC}$}
\newcommand{\NOISE}{$\mathtt{Noise}$}
\newcommand{\ENSEMBLE}{$\mathtt{Ensemble}$}
\newcommand{\Plus}{\texttt{+}}

\usepackage[urlcolor=blue,bookmarks=false,hypertexnames=true,citecolor=blue]{hyperref}

\title{\LARGE \bf
Grasping with Chopsticks: Combating Covariate Shift in Model-free Imitation Learning for Fine Manipulation}

\author{Liyiming Ke, Jingqiang Wang, Tapomayukh Bhattacharjee, Byron Boots and Siddhartha Srinivasa
\thanks{University of Washington, Seattle WA 98105 USA.
        {\tt\small \{kayke, jwq123, tapo, bboots, siddh\}@uw.edu}}%
}

\begin{document}

\maketitle
\thispagestyle{empty}
\pagestyle{empty}

\input{inputs/0-abstract}

\input{inputs/1-intro}

\input{inputs/2-method}

\input{inputs/3-problem}

\input{inputs/4-result}

\input{inputs/5-discussion}

\addtolength{\textheight}{-6cm}  


\input{inputs/777-appendix}

\input{inputs/888-acknowledgement}



\bibliographystyle{IEEEtran}
\bibliography{main}

\end{document}

%% file: inputs/999-inline-notes.tex
\newboolean{include-notes}
\setboolean{include-notes}{true}

\newcommand{\kknote}[1]{\ifthenelse{\boolean{include-notes}}%
 {\textcolor{orange}{\textbf{K: #1}}}{}}
\newcommand{\tbnote}[1]{\ifthenelse{\boolean{include-notes}}%
 {\textcolor{cyan}{\textbf{Tapo: #1}}}{}}
\newcommand{\ssnote}[1]{\ifthenelse{\boolean{include-notes}}
 {\textcolor{red}{\textbf{SS: #1}}}{}}
\newcommand{\bbnote}[1]{\ifthenelse{\boolean{include-notes}}
 {\textcolor{green}{\textbf{BB: #1}}}{}}

%% file: inputs/0-abstract.tex
\begin{abstract}

Billions of people use chopsticks, a simple yet versatile tool, for fine manipulation of everyday objects. The small, curved, and slippery tips of chopsticks pose a challenge for picking up small objects, making them a suitably complex test case. This paper leverages human demonstrations to develop an autonomous chopsticks-equipped robotic manipulator. 
Due to the lack of accurate models for fine manipulation, we explore model-free imitation learning, which traditionally suffers from the \emph{covariate shift} phenomenon that causes poor generalization. We propose two approaches to reduce covariate shift, neither of which requires access to an interactive expert or a model, unlike previous approaches. First, we alleviate single-step prediction errors by applying an invariant operator to increase the data support at critical steps for grasping. Second, we generate synthetic corrective labels by adding bounded noise and combining parametric and non-parametric methods to prevent error accumulation. We demonstrate our methods on a real chopstick-equipped robot that we built, and observe the agent's success rate increase from $37.3\%$ to $80\%$, which is comparable to the human expert performance of $82.6\%$.

\end{abstract}

%% file: inputs/1-intro.tex
\section{Introduction}

\begin{figure}[t]
  \centering
  \includegraphics[width=\linewidth]{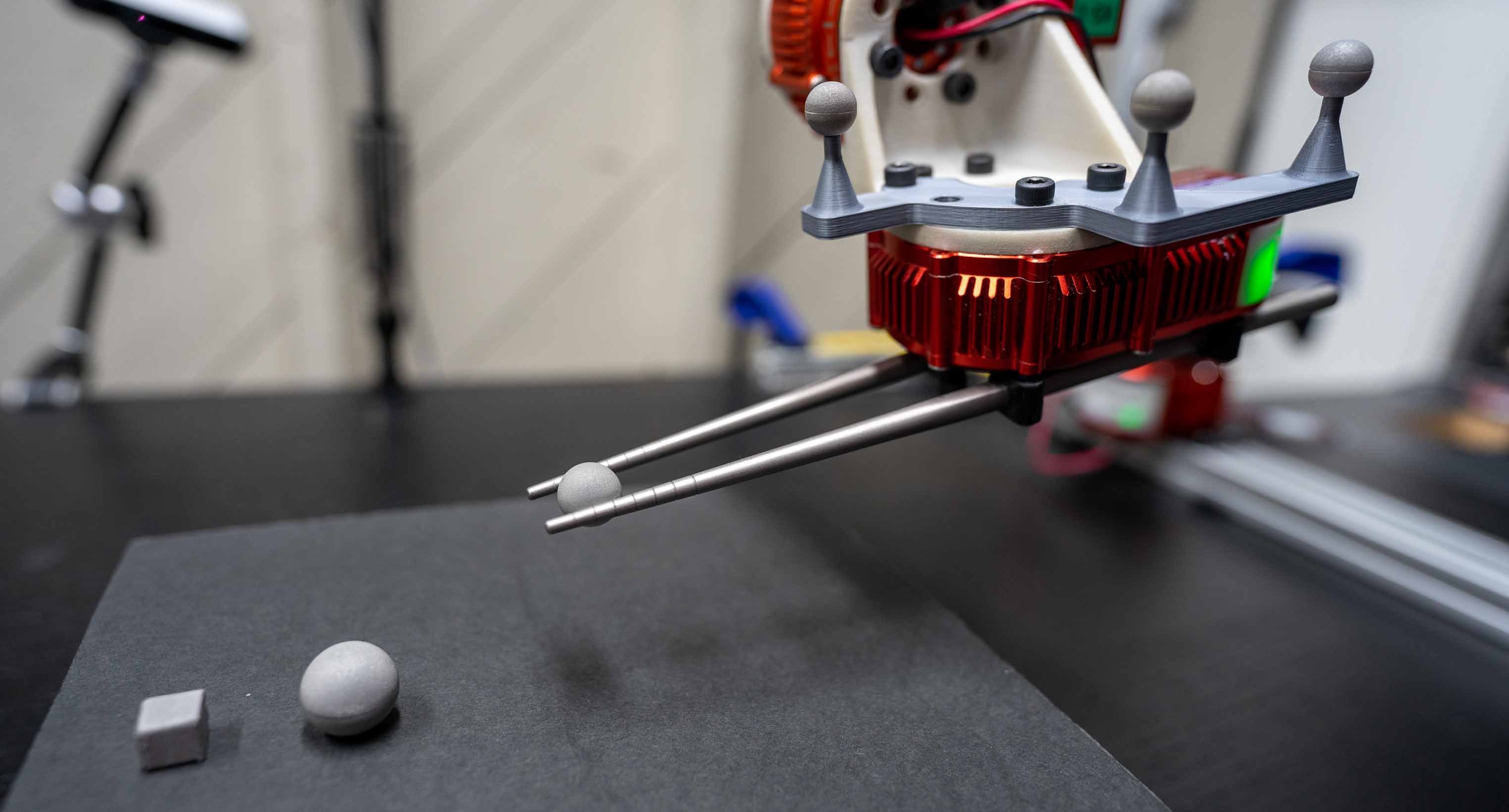}
  \caption{Fine manipulation using chopsticks.}
  \label{fig:system_task}
  \vspace{-1em}
\end{figure}

Although complex end effectors are inherently suited to fine manipulation~\cite{finemanip} due to fewer design constraints, simple tools are easier to study and deploy, and they are ubiquitous in industrial manipulators. With a \emph{human-in-the-loop}, simple end effectors can also perform general fine manipulation~\cite{marohn2004davinci, kent2017comparison}. We choose chopsticks, a simple tool that is very familiar to humans, as an example to learn and automate fine manipulation strategies from human demonstrations. To that end, we have built an automated chopstick-equipped robot comprised of a 6DOF robot arm outfitted with a 1DOF actuated chopstick (Fig.~\ref{fig:system_task}). Our goal is to demonstrate autonomous superhuman chopstick dexterity with our robot.

The efficacy of chopsticks' design has inspired researchers to adapt them for diverse robotic applications, such as surgery~\cite{sakurai2016thin,joseph2010chopstick,ragupathi2010robotic}, micro-manipulation~\cite{ramadan2009developmental}, and meal assistance~\cite{chang2007pincer,yamazaki2012autonomous}. However, chopsticks' practicality and generality in applications come at the cost of complexity to control~\cite{mason2012autonomous}. Their small, curved, and slippery tips require precise movements for grasping small and rigid objects such as a toy marble. Their limited allowance for failures makes them a suitably complex test case for evaluating fine manipulation tasks. Noticeably, humans have demonstrated impressive adaptability in teleoperating a robot equipped with chopsticks to pick up hard-to-grasp small objects~\cite{ke2020telemanipulation}. We aim to leverage human demonstrations to learn control policies using \emph{imitation learning}~\cite{billard2011imitation,mulling2013learning}. The challenge we face is further exacerbated by the lack of accurate models for our assembled robotic test-bed~\cite{Robotics_undated-vk}, a common constraint for fine manipulation tasks~\cite{cutkosky2012robotic,billard2019trends}. 

The lack of accurate models motivates our study of model-free imitation learning~\cite{ho2016model,pastor2009learning,dyrstad2018teaching,zhang2018deep}. Here, we have access to demonstration data but not to the expert's policy function or the environment's transition model. Under these conditions, supervised learning methods like \emph{behavior cloning}~\cite{pomerleau1989alvinn} learn a policy function by matching the expert's action distribution. Minimizing action distribution divergence, however, does not necessarily guarantee the recovery of parsimonious states that lead to task success~\cite{osa2018algorithmic}. A learned agent can suffer from \emph{covariate shift}~\cite{quionero2009dataset}, i.e., compounding errors in the action space that lead the agent to unseen states during test time. This problem can be especially detrimental for fine manipulation, the success of which critically depends on a few steps that usually occur near the end of a trajectory. 

To remedy covariate shift, researchers have proposed \emph{interactive} imitation learning methods, such as DAgger~\cite{ross2011reduction} and DART~\cite{laskey2017dart}, to query an expert \emph{online} for corrective labels. DAgger rolls out a learned agent and asks the expert for labels on learner visited states, which can be computationally expensive and unnatural on a teleoperation interface~\cite{laskey2017dart}. DART injects noise during data collection, disturbs expert teleoperation, and forces the expert to provide corrective labels. However, injecting noise during data collection can burden the expert: adding a small amount of random noise for our fine manipulation task, as DART suggests, would require the expert to spend $43\%$ more time on collecting data.\footnote{Though $95\%$ of the noise injected resulted in at most $0.35^\circ$ deviation per joint, it lowered the expert success rate by $18\%$ and forced the expert to spend more time completing each trajectory.}

These challenges prompt us to address covariate-shift in model-free imitation learning in a \emph{non-interactive} setting, where we have access to demonstration data but not to an interactive expert. Since covariate shift results from the interplay of single-step errors and their accumulation over time, our key ideas are to (1) increase data support to address single-step errors, and (2) provide corrective labels to address the accumulation of errors. Specifically, we provide:
\begin{itemize}
\item \emph{Enhanced data support} by transforming the data to an object-centric frame that preserves the relative transformation between the end effector and object, while making training data denser around the \emph{critical} region for grasp success.
\item \emph{Corrective labels by injecting noise} into the collected state, assuming the same action may serve as the \emph{corrective label} for the deviated state. Thus, we implicitly enforce smoothness to the learned policy and tell the agent how to recover from deviated states.
\item \emph{Corrective labels by choosing a combination of parametric and non-parametric methods} that improve matching of the action distribution at unseen states. Because of our problem structure, a better match in action distribution leads to a higher likelihood of matching the state distribution, preventing error accumulation. 
\end{itemize}

We demonstrate our proposal's effectiveness on a physical robot equipped with chopsticks to pick up small cube- and ball-shaped objects, as shown in Fig.~\ref{fig:system_task}. Our proposed agent achieves 60\% success rates picking up even the most challenging item, a small ball, whereas a naive behavior cloning agent has only a 12\% success rate. Our agent achieves an 80\% average success rate picking up all three objects tested, comparable to the expert human performance of 82\%. We conduct ablation tests, visualize the resulting states' distribution, and observe a smaller covariate shift from our proposed agents. We also validate the generality of the noise injection method on several Mujoco simulated tasks.

Our promising empirical results, based on pragmatic assumptions of data support and policy smoothness, open the door for further theoretical analysis of combating covariate shift. Furthermore, although we have focused on the \emph{non-interactive} setting, our techniques directly transfer to the \emph{interactive} setting, enhancing robustness while reducing user burden.

%% file: inputs/2-method.tex
\section{Methods}\label{sec:method}

\subsection{Transform: Increasing Data Support}
\label{ssec:obj_centric}

Our goal is to develop an agent that can generalize from demonstration data to predict an action for any query state. However, we lack data support for some states (e.g., the ``unseen state'' during rollout). We propose to apply an invariant operator to transform the data, making it denser around the region of interest and thus increasing the data support.

\begin{figure}[!t]
\vspace{.5em}
\begin{subfigure}{.49\linewidth}
  \centering
  \includegraphics[width=\textwidth]{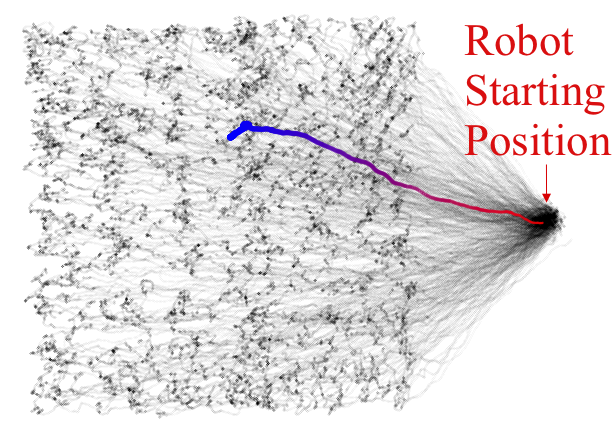}
  \caption{Robot-centric frame.}
  \label{fig:frame_robot}
\end{subfigure}
\begin{subfigure}{.49\linewidth}
  \centering
  \includegraphics[width=\textwidth]{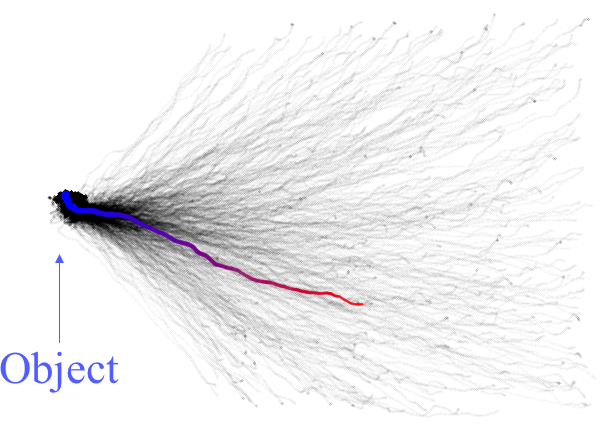}
  \caption{Object-centric frame.}
  \label{fig:frame_obj}
\end{subfigure}
\caption{Visualizing the end-effector positions for all demonstrations under different coordinate frames. Each black dot is an xyz-position of the end effector in one step. We highlight one trajectory, which starts with red dots and ends in blue.}
\label{fig:frame}
\end{figure}

In manipulation, changing the frame of reference can significantly change the distribution of trajectories (Fig.~\ref{fig:frame}). We could choose a \emph{robot-centric} frame, where the robot base is the origin, or an \emph{object-centric} frame~\cite{mason2011generality}, where the object location is the origin. We propose that using an object-centric frame can reduce the covariate shift and improve the policy generalization, especially for fine manipulation. The transformation to an object-centric frame would result in a denser distribution of trajectories near the origin where the object is located, increasing data support for this critical region that determines grasping success. Using an object-centric frame also allows the policy learned to be invariant to the translation of object location. This makes the learned policy more sample efficient when generalizing to novel object locations.

\subsection{Noise: Generating Synthetic Corrective Labels}

\begin{figure*}[!htb]
\begin{subfigure}[b]{.3\linewidth}
  \centering
  \includegraphics[width=\textwidth]{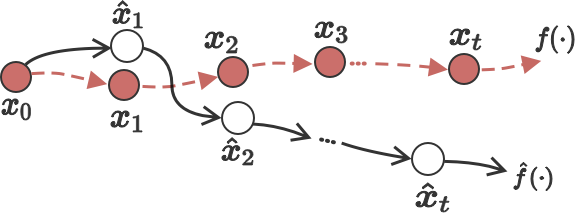}
  \caption{Covariate shift: A learner roll out (black) deviates from the demonstration (red) and error accumulates.}
  \label{fig:cov_shift}
\end{subfigure}
\hspace{1em}%
\begin{subfigure}[b]{.3\linewidth}
  \centering
  \includegraphics[width=\textwidth]{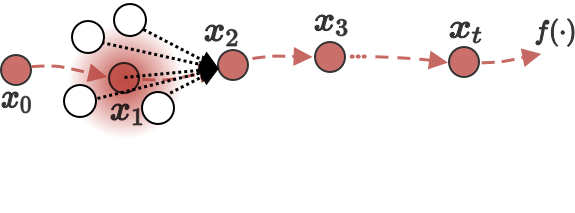}
  \caption{Inject noise into the collected states and reuse the collected action as synthetic corrective labels.}
  \label{fig:noise_gen}
\end{subfigure}
\hspace{1em}%
\begin{subfigure}[b]{.3\linewidth}
  \centering
  \includegraphics[width=\textwidth]{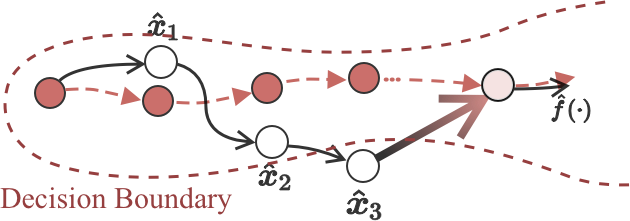}
  \caption{Use non-parametric methods (like a k-NN) to return the agent to proper region when deviations occur.}
  \label{fig:ensemble}
\end{subfigure}
\caption{Prevent error escalation in imitation learning.}
\label{fig:corrective_labels}
\end{figure*}


Although the transformation technique we use improves the agent's success rate, we still observe significant deviations during test time that result in task failure (Fig.~\ref{fig:cov_shift}). This is understandable because machine learning algorithms generally need exponentially more data for progressive improvement~\cite{laskey2017dart}. Instead of naively collecting more data, we introduce corrective action labels that can help the agent recover from deviations. For example, Venkatraman \textit{et al.}~\cite{venkatraman2015improving} rolled out trained agents, collected their deviation states and used model-predictive control to generate corrective labels to go back to the demonstrated trajectory. Unfortunately, models sufficiently accurate for fine manipulation can be challenging to build.

We propose to generate \emph{synthetic} corrective labels by injecting noise into the collected demonstration \emph{states} (``deviated state'') and reusing the collected action (``corrective labels''), thus not requiring access to an expert or a model. Unlike DART and DAgger, which emphasize collecting corrective labels for the states that the agent will visit during rollout (test state distribution), we hypothesize that we do not need to match the deviated states' distribution accurately. Instead, we need to collect enough corrective labels to \emph{cover} the deviated states' distribution. Since we can generate labels for free without burdening an expert, we choose to generate labels for randomly sampled deviated states, thus simplifying the selection of states for which to generate synthetic corrective labels. Fig.~\ref{fig:noise_gen} shows an example where we sample states around a demonstrated state and reuse the demonstrated action as synthetic corrective labels.

Researchers have injected noise~\cite{bishop1995training} into problems that reduce a high-dimensional input to a low-dimensional output, e.g., for classification~\cite{sietsma1991creating} and object recognition in visual and language domains~\cite{shorten2019survey}. In these works, such tasks are invariant under a wide variety of transformations~\cite{goodfellow2016deep}. However, our robotic manipulation task has \emph{low-dimensional} states and actions, where the mapping learned may not be invariant to the noise. We provide two insights to justify why injecting noise can still be desirable.

First, we apply a \emph{small} amount of additive Gaussian noise to the demonstration state instead of a large amount that could pollute the data by mapping a state to a detrimental action. Inspired by~\cite{poole2014analyzing}, which showed the effectiveness of noise injection for autoencoders by carefully tuning the magnitude of the noise, we generate Gaussian noise $\epsilon \sim \mathcal{N}(0, \sigma)$ to add to the collected states, where $\sigma$ is the covariance of the noise. For simplicity, we correlate the noise size $\sigma$ with the variance of the data.

Second, because of the structure of our problem, the collected action can serve as the corrective label for the noise-injected deviated state. Our state and action representations both include the end-effector pose. Therefore, when an agent starts drifting from a demonstrated trajectory and enters a deviated state, our algorithm can teach it to return to the original trajectory by reusing the same action label. Injecting noise can also ensure the learned policy is smooth, which is desired since we assume the actions are Lipschitz continuous w.r.t the states.

\subsection{Ensemble: Following the Expert Advice}

We can reduce error accumulation at unseen states by choosing methods that more effectively match the action distribution. A neural network's optimization objective is limited to its training data and will not necessarily generalize well to unseen inputs~\cite{russell2002artificial}. In contrast, non-parametric methods generate test outputs by combining the training data, their predictions must come from the training data and are therefore constrained~\cite{altman1992introduction}. For example, a k-nearest neighbor (\KNN) agent will not cause the robot to move its joint positions beyond the interpolation of its training data. 

We use \KNN~in conjunction with behavior cloning (\BC). Specifically, our agent follows the \KNN~predicted action if the query state deviates from the training data (Fig.~\ref{fig:ensemble}).

By using the \KNN~method, we are \emph{forcing a known action} to a new unseen state during test time to ensure the action distribution during training and testing will match. For our manipulation task, the state and action both include the robot's end-effector pose. Sending a \emph{known action} is equivalent to sending the agent to \emph{a known state}, implicitly reducing the agent's deviation from training data, thus reducing covariate shift. However, nonparametric method's performance is subject to its distance function, which can be difficult to design for high-dimensional data.

The distance function for non-parametric methods serves two purposes: (1) to evaluate the proximity of a query to the stored data points; and (2) to weight and combine the expert labels. Our key observation is that (1) requires only a rough estimate of the distance to decide whether a query state is far from the training data, and (2) needs a carefully tuned distance function to assign weights to expert labels. Therefore, we propose to use a decision tree to invoke a \KNN~agent \emph{only} when the distance of the query state is far from its nearest neighbors and invoke a behavior cloning neural network agent otherwise. By invoking \KNN~only when the agent is far away, we bypass the need to carefully design a distance function for it, favor \BC's scalability with data when we are inside the training data distribution, and rely on \KNN~to correct the agent's deviation when we are outside the training data distribution.


%% file: inputs/3-problem.tex
\section{Experiments}

\begin{figure*}
\vspace{.6em}
\begin{subfigure}{.32\linewidth}
  \centering
  \includegraphics[width=\linewidth]{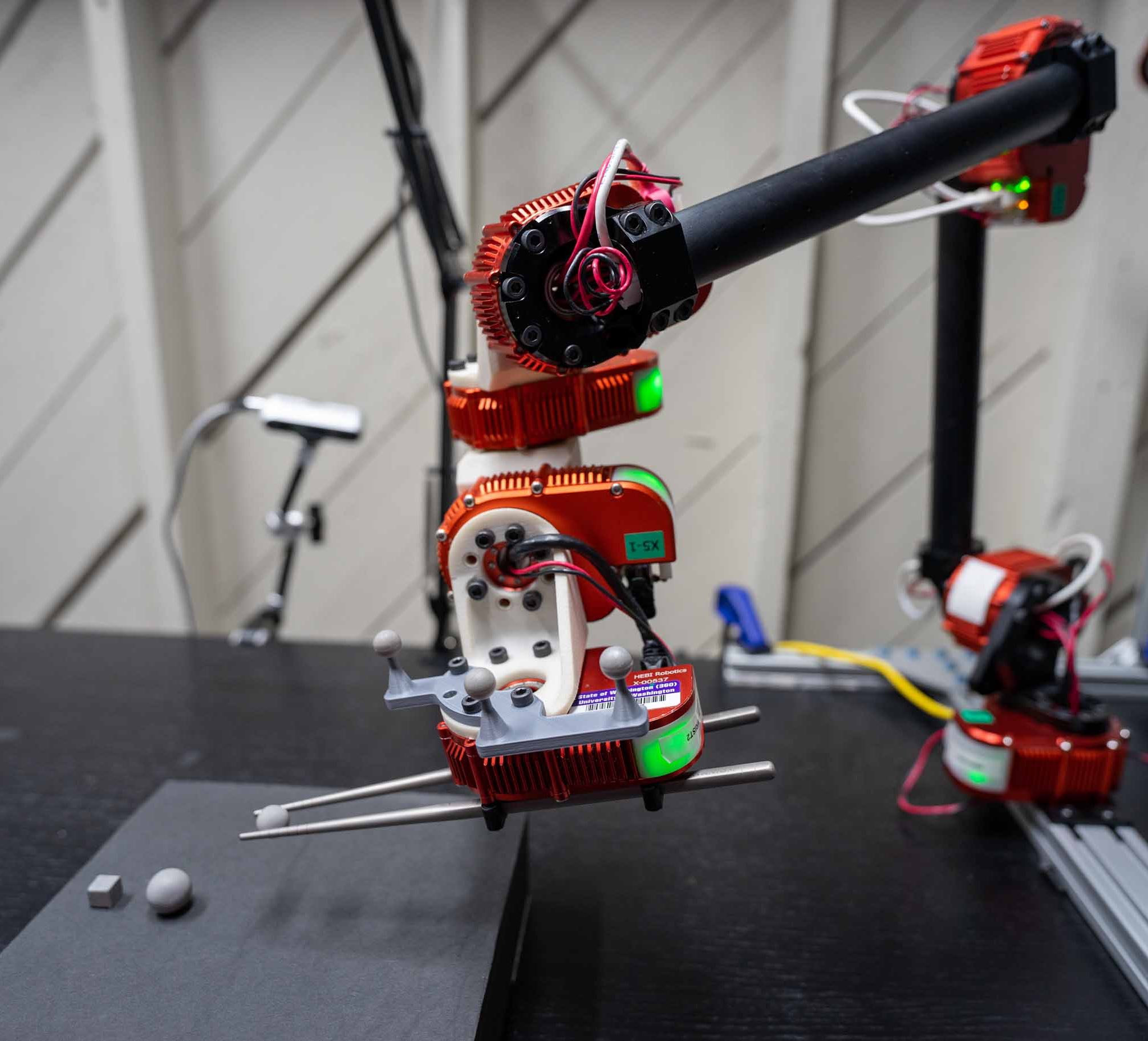}
  \caption{Robot platform.}
  \label{fig:problem_robot}
\end{subfigure}
\begin{subfigure}{.32\linewidth}
  \centering
  \includegraphics[width=\linewidth]{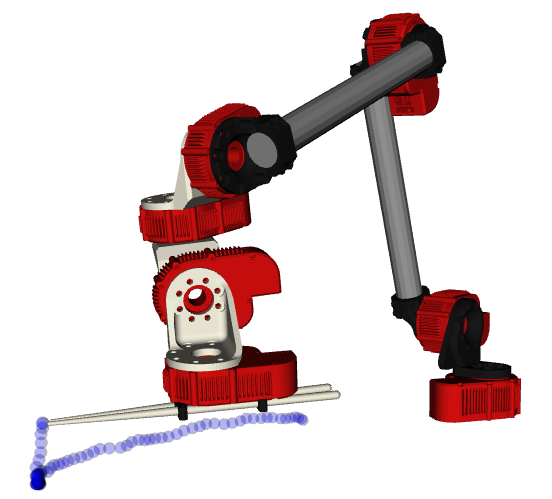}
  \caption{Example demonstration.}
  \label{fig:problem_traj}
\end{subfigure}
\begin{subfigure}{.32\linewidth}
  \centering
  \includegraphics[width=\linewidth]{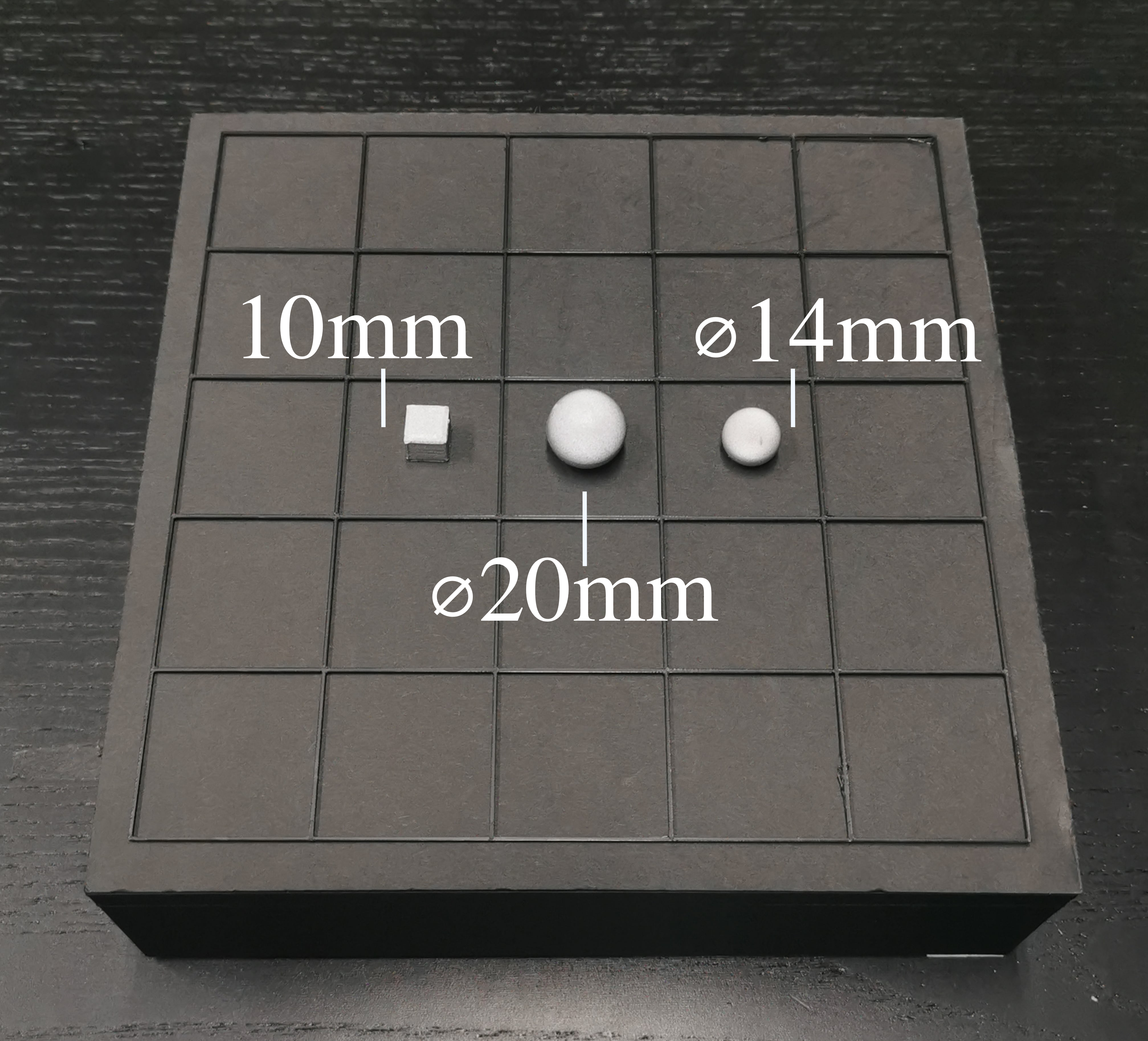}
  \caption{Evaluation.}
  \label{fig:problem_obj}
\end{subfigure}
\label{fig:system}
\caption{Experiment setup.}
\end{figure*}

\subsection{Experimental Setup}

We built a 6-DOF robot (Fig.~\ref{fig:problem_robot}) equipped with a pair of chopsticks as its end effector in order to develop algorithms that control the chopsticks to pick up challenging objects: a cube with a $1$\,cm edge length, a ball with a $2$\,cm diameter, and another ball with $1.4$\,cm diameter, as shown in Fig.~\ref{fig:problem_obj}. The kinematic model for our inexpensive hardware is not highly accurate since the robot is assembled from parts with joints that are not strictly rigid. Even with the best calibration, inaccuracies still accumulate along robot links and result in position errors ranging from 1 mm to 6 mm at the robot's end effector. This implies that the difference between the calculated chopstick tip position and its actual position is comparable to the radius of the small objects used in our experiments. For each object, we collected $500$ trajectories from an expert teleoperating the robot to pick up the object (Fig.~\ref{fig:problem_traj}). The data collection setup is the same as in our previous work~\cite{ke2020telemanipulation}.

Our agent had access to the tracked location of the objects and the robot's end-effector pose. We defined success as \emph{grasping} the objects using chopsticks, \emph{lifting} them above the workstation, and \emph{holding} them in the air for $1$\,s. We evaluated the performance of each method on each object by computing the success rate over 25 trials. During evaluation, we divided the square workstation plate into a $5 \times 5$ grid (Fig.~\ref{fig:problem_obj}) and placed the object in the center of each grid cell to ensure effective coverage over the entire workspace. See Appendix~\ref{app:experiment} for more details.

\subsection{Experimental Procedure}
We compared our methods in Section~\ref{sec:method} with human demonstrations during teleoperation ($\mathtt{Expert}$) and a replay of the successful demonstrations ($\mathtt{Replay}$). $\mathtt{Replay}$ tests the \emph{repeatability} of our hardware. We chose successful demonstrations, placed objects at \emph{exactly the same locations} used during data collection, and replayed the demonstrations to see if the robot could pick up the objects.

We used two baselines. The first is a parametric method, \BC\Plus\ROBOTC, a neural-network based behavior cloning agent that uses the default robot-centric frame. The second is a non-parametric method, \KNN\Plus\ROBOTC, which is a k-nearest-neighbor agent that also uses the robot-centric frame.

We evaluated three methods as described in Section~\ref{sec:method}: (1) using the object-centric frame to train behavior cloning and the k-nearest neighbors agents, \BC\Plus\OBJC~and \KNN\Plus\OBJC, respectively, (2) injecting a small amount of Gaussian noise into the behavior cloning agent, \BC\Plus\OBJC\Plus\NOISE, and (3) combining the parametric method \BC\Plus\OBJC\Plus\NOISE~and non-parametric method \KNN\Plus\OBJC~via a decision tree model, denoted as \ENSEMBLE. Implementation details are shown in the Appendix~\ref{app:details}.

%% file: inputs/4-result.tex
\section{Results}

\begin{table}
\begin{center}
\begin{tabular}{l|c|c|c|c}
\toprule
 Method & Cube & Ball$\diameter 20$\,mm & Ball$\diameter 14$\,mm & All \\
\midrule 
$\mathtt{Expert}$   & 100  & 80        & 68    & 82.7    \\
$\mathtt{Replay}$   & 100  & 80        & 80    & 86.7    \\
\midrule
\BC\ \Plus\ROBOTC & 84   & 16        & 12    & 37.3    \\
\BC\ \Plus\OBJC  & 92   & 16        & 24 & 44.0        \\
\BC\ \Plus\OBJC\Plus\NOISE & 92 & 76 & 48 & 72.0 \\
\midrule
\KNN\ \Plus\ROBOTC & 64   & 28        & 8     & 33.3    \\
\KNN\ \Plus\OBJC  & 84   & 64        & 12 & 53.3       \\
\midrule
\rowcolor{Gray}
\ENSEMBLE & \textcolor{blue}{\textbf{96}} & \textcolor{blue}{\textbf{84}} & \textcolor{blue}{\textbf{60}} & \textcolor{blue}{\textbf{80.0}} \\
\bottomrule
\end{tabular}
\end{center}
\caption{Percentage success rates evaluated over 25 trials.}
\label{fig:grasping_results}
\end{table}

\begin{figure*}[]
\vspace{.6em} 
\begin{subfigure}{.32\linewidth}
  \centering
  \includegraphics[width=.8\linewidth]{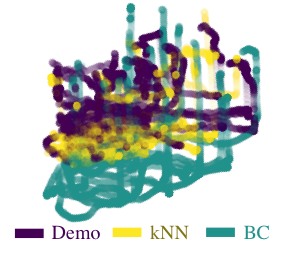}
  \caption{\KNN\Plus\OBJC~versus \BC\Plus\OBJC.}
  \label{fig:pca_bc_vs_knn}
\end{subfigure}
\begin{subfigure}{.32\linewidth}
  \centering
  \includegraphics[width=.8\linewidth]{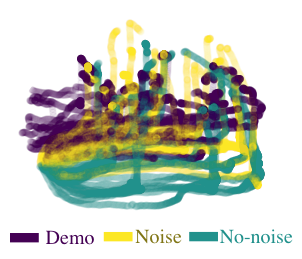}
  \caption{\BC\Plus\OBJC~versus \BC\Plus\OBJC\Plus\NOISE.}
  \label{fig:pca_if_noise}
\end{subfigure}
\begin{subfigure}{.32\linewidth}
  \centering
  \includegraphics[width=.8\linewidth]{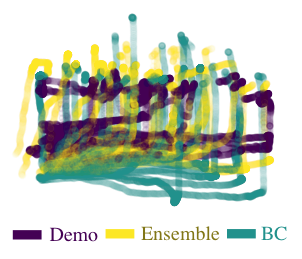}
  \caption{\ENSEMBLE~versus \BC\Plus\OBJC\Plus\NOISE.}
  \label{fig:pca_ensemble}
\end{subfigure}
\caption{Comparing the state distributions as a proxy of covariate shift for trained agents. Each dot is a state in the agent's roll out. Green states are farther away from the demonstrations (purple), indicating that their corresponding agent suffers from more covariate shift than the yellow agent.}
\label{fig:cov_shift_viz}
\end{figure*}

\subsection{Success Rates for Fine Manipulation}

The experimental results are shown in Table.~\ref{fig:grasping_results}, and the best performers in each column are highlighted. Our parametric method baseline, \BC\Plus\ROBOTC, and nonparametric method baseline, \KNN\Plus\ROBOTC, had relatively low success rates. However, the causes of their failures differ. \BC\Plus\ROBOTC~has difficulty picking up objects that are placed farther away from the robot. The agent tends to reach towards the wrong location after moving over a long distance to approach the object, highlighting the covariate shift's impact. In contrast, the \KNN\Plus\ROBOTC~agent's poses look more similar to expert demonstrations. However, its trajectories are not smooth and sometimes end abruptly on top of the object without picking it up. This occurs because \KNN~does not guarantee a smooth policy function; even after careful tuning of the distance function, it was challenging to eliminate the jerky motions. \KNN's sudden stops are due to direct imitation of the training data. During demonstration, the human expert often slowed or even paused their movements around the object, adjusting the approaching pose before closing the chopsticks and lifting the object. The distance function we chose fails to select and mix the more relevant action labels. This confirms the sensitivity of \KNN~to its distance function. 

Transforming to the \OBJC~frame improved the success rates for \KNN~and \BC~by $20$\% and $6.7$\%, respectively. \KNN~becomes less likely to generate jerky motions or stop since it benefits from the increased data support. \BC~still suffers from covariate shift, but the agent has a higher likelihood of reaching towards the object due to denser data distribution near it.

Injecting noise to \BC\Plus\OBJC~during training increases its success rate by $28\%$. When the items are \emph{close} to the robot, the agent has an almost $100\%$ success rate picking up even the most challenging item. For objects that are far away, the robot sometimes picks up the object by successfully reaching the location; at other time, it ends up merely rotating the chopsticks. 

Using a decision tree to combine our best parametric method (\BC\Plus\OBJC\Plus\NOISE) and non-parametric method (\KNN\Plus\OBJC) yields the highest performing agent that achieves near-expert performance. During test time, if a state's distance to its nearest neighbors exceeds a threshold, the agent triggers the non-parametric method to bring the state back. We observe that almost all rollouts trigger the non-parametric method at least once. No matter how far an object is placed, the \ENSEMBLE~agent can reach it in a ``standard'' way that is similar to the pose demonstrated by the expert. Failures occasionally occur as the agent misses the grasping point by some sub-mm error.

\subsection{Covariate Shift Across Methods}

To gauge the covariate shift for different agents, we visualize the distributions of their test states. We collect 25 rollouts from each agent, record the robot-visited states, and plot the state distribution after dimensionality reduction using Principal Component Analysis (PCA), as shown in Fig.~\ref{fig:cov_shift_viz}. First, we observe that \BC~encounters more covariate shift than \KNN, i.e., that states visited by \KNN~are closer to the demonstrated states, confirming that a better matching of action distribution will lead to a better match of state distribution. Second, injecting noise into \BC~results in less covariate shift than no noise, verifying that noise injection can provide effective correcting labels. Third, the \ENSEMBLE~model that combines \BC~with \KNN~has less covariate shift than using \BC~alone.

\subsection{Noise Injection: Validation through MuJoCo Environments}

\begin{figure}[t]
  \centering
  \includegraphics[width=.92\linewidth]{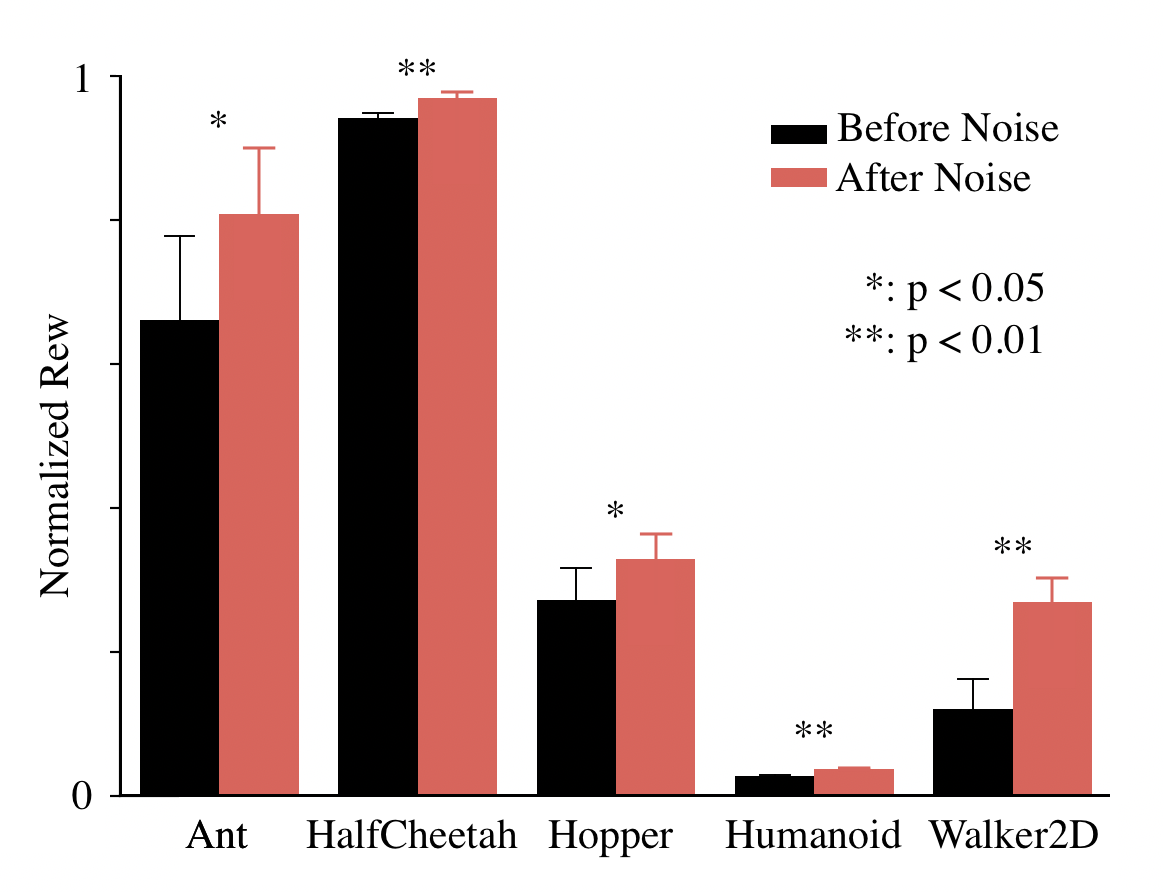}
  \caption{Performance comparison before and after noise injection. For each MuJoCo environment and each condition, we trained 5 agents using consecutive random seeds. The performance difference is statistically significant under paired T-tests.}
  \label{fig:mujoco}
\end{figure}

We apply the noise injection method to MuJoCo simulated environments~\cite{todorov2012mujoco} to test the method's generality. We use demonstration data from~\cite{ucb2020drlclass}, train $5$ behavior cloning agents under consecutive random seeds as baselines, and train another $5$ agents with noise injection for comparison. Figure~\ref{fig:mujoco} compares performances before and after noise injection. A paired T-test shows that $p < 0.05$ for all environments. There is strong evidence that, on average, noise injection improves the imitation learner.

Though the performance gains for some simulated environments are not as significant as those we see for our real robot, we think the difference may be due to the demonstration source. We use ``real'' human data for our real robot experiment versus the ``synthetic expert demonstration'' generated by a reinforcement learning (RL) agent for the MuJoCo tasks. Human experts are known to exhibit multi-modal behaviors during demonstrations, whereas trained RL agents tend to have single modes in their reaction~\cite{ke2019imitation}. Given that noise injection improves the success rate for our physical robot task by a considerable $28$\%, further inquiry is needed to determine if noise injection is better at enhancing learning from multi-modal demonstration data.

%% file: inputs/5-discussion.tex
\section{Discussion}

We leave some topics for future work. During noise injection, for simplicity, we experiment only with independent multivariate Gaussian noise with a ﬁxed size of covariance. It is worth exploring how to formalize the bounded noise and analyzing how different task domains may beneﬁt from different noise shapes. For the ensemble model, future work could explore an alternative way to switch between \KNN~and \BC~agents in the \ENSEMBLE~model, perhaps by learning a threshold condition from the data. 

Our work critically depend on two key assumptions. First, to increase data support by applying an invariant operator, we assume the existence of a \emph{critical} region that demands more data support. Second, to reuse collected action labels and leverage a nonparametric method to generate corrective labels, a more accurate match of action distribution should lead to a more accurate match of the states. The assumption holds if a part of the state and action representation is directly connected, e.g., the robot state contains its joint position, and the robot command accepts the target joint position. The assumption does not hold, for example, if the robot is torque-controlled; in these cases, further exploration on how a learner can generate synthetic corrective labels is needed.

Nevertheless, our proposals do not assume access to a model or an interactive expert and are therefore more easily applicable to fine manipulation tasks. Compared to DAgger and DART, which collect corrective labels from experts, we can generate synthetic corrective labels for free. Because of the relatively lower cost of doing so, we generate labels for randomly sampled state distributions that \emph{cover} the deviated state distribution without accurately \emph{matching} it. Though our proposals focus on a non-interactive setting, they can directly transfer to an interactive one. 

We choose model-free imitation learning because an \emph{accurate} model for fine manipulation is rare. However, it remains to be seen how to leverage an \emph{inaccurate} model in imitation learning. This work is but a first step towards exploring general-purpose autonomous fine manipulation using simple tools. We look forward to extending it by combining model-free and model-based methods to manipulate a more diverse set of hard-to-grasp small objects.

%% file: inputs/777-appendix.tex
\section*{Appendix}

\subsection{Experimental Setup}
\label{app:experiment}
\paragraph{Robotic Testbed.}
We use the end-effector (EE) pose to describe the robot's state, which is an 8D vector containing (1) the x-y-z position of the bottom chopstick tip, (2) a quarternion representation of the rotation of the chopsticks, and (3) the opening angle of the last joint. We command the robot by sending a target end-effector pose at 100Hz, using an Inverse Kinematics solver to translate to joint positions and running a PID controller at 500Hz to move each joint.

\paragraph{Calibration Improved Performance.}
The default model and controller for our hardware were not highly accurate. The average EE position error was $10$\,mm. After careful calibration, we reduced this error to $4$\,mm. Initially, even a well-tuned controller had low success rates for picking up a cube and small ball during replay (90\% and 15\%, respectively). We implemented a custom PID controller and gain-tuning to achieve 100\% and 80\%, respectively.

\paragraph{Demonstrations.}
We collect the demonstration at 100 Hz to match the test scenario. Each trajectory contains an average of 600 (state, action) pairs. The state is a 11-D vector containing the robot's state and the object's x-y-z position. The action is the target end-effector pose. During each trajectory, we initiate the robot around a fixed home configuration and place the object at a random location across the workstation. One expert user collect all trajectories to reduce multi-modal behavior that might interfere with learning (e.g., picking up object using different strategies). We filter out failed trajectories and keep only the 500 successful ones. 

\subsection{Implementation Details}
\label{app:details}
\paragraph{\BC.}
We trained a two-layer fully connected neural network of size $64 \times 32$ with $ReLU$ activation. It outputs the 8D target end0effector pose. To compute its loss, we divided the 8D pose to position, rotation, and opening angle and computed the loss for each component using the mean squared error or the rotation difference. We then used a weighted linear combination to sum the components' losses to a 1D loss. The weights are tunable parameters.



\paragraph{\KNN.}
We used the last $3$ end-effector poses and the current object location as input to the \KNN~agent. We specify its distance function to be similar to the \BC~loss function but use a different set of weights.

\paragraph{\NOISE.}
During training of \BC~ agents, instead of optimizing $\sum_{i \in \text{ batch}}[f(x_i) - a_i]$, where $x_i$ is the state and $a_i$ is the action, we sample 20\% of the data in each batch and replace the state $x_i$ with $x'_i = x_i + \epsilon$, where $\epsilon \sim \mathcal{N}(0, \sigma)$. $\sigma$ is a diagonal matrix whose diagnoal entries are $\eta\hat{\sigma}$.  We choose a fixed noise magnitude, $\eta$, for all dimensions of the state and $\hat{\sigma}$ is the variance of each dimension of the state. Empirically, $\sigma = \eta$ also achieves comparable performance.

\paragraph{\ENSEMBLE.}
Given that \KNN~yields the nearest neighbors for a state and the corresponding distances, we set a threshold parameter $\alpha$ such that the agent follows \BC~ iff $\sum(d_i) / k < \alpha$, where $d_i$ the distance to the i-th closest neighbor. Further details are in~\cite{supplementary}.

%% file: inputs/888-acknowledgement.tex
\section*{Acknowledgement}
Research reported in this publication was supported by the Eunice Kennedy Shriver National Institute Of Child Health \& Human Development of the National Institutes of Health under Award Number F32HD101192. The content is solely the responsibility of the authors and does not necessarily represent the official views of the National Institutes of Health. This work was also (partially) funded by the National Science Foundation IIS (\#2007011), National Science Foundation DMS (\#1839371), the Office of Naval Research, US Army Research Laboratory CCDC, Amazon, and Honda Research Institute USA.